\documentclass[runningheads]{llncs}
\usepackage{eccv}

\usepackage{eccvabbrv}
\usepackage{graphicx}
\usepackage{booktabs}
\usepackage{xcolor}
\usepackage{url}

\usepackage[accsupp]{axessibility}  
\usepackage{hyperref}
\usepackage{orcidlink}

\begin{document}

\title{Caption Bottleneck Models} 

\author{Seref Baris Cagliyan\inst{1}\orcidlink{0009-0003-5370-7915} \and
Umut Ozdemir\inst{1}\orcidlink{0009-0007-7259-9007} \and
Merve Tapli\inst{1}\orcidlink{0000-0002-6161-6488} \and
Emre Akbas\inst{1, 2}\orcidlink{0000-0002-3760-6722}}

\authorrunning{B.~Cagliyan et al.}

\institute{Dept. of Computer Eng., Middle East Technical University (METU) \\
 \and
Robotics \& AI Center (ROMER), METU}

\maketitle

\begin{abstract}
Concept Bottleneck Models (CBMs) provide interpretability by routing predictions through a layer of human-understandable concepts. 
However, defining an optimal concept set for a specific dataset remains an open challenge. 
Existing approaches rely on expensive expert annotations or LLM-generated lists based solely on class names. 
Even ``open-vocabulary'' variants typically depend on static concept sets, which restrict discovery and introduce label bias. 
Furthermore, traditional CBMs often suffer from information leakage, where unmodeled visual features bypass the bottleneck and compromise the integrity of the explanations. 
To overcome these limitations, we propose Caption Bottleneck Models (CaBM), a framework that circumvents the need for predefined concept sets by replacing rigid concept layers with free-form natural language. 
By representing images via LMM-generated captions and training a classifier strictly on this text, CaBM ensures a leakage-free architecture by construction.
Additionally, by analyzing the text classifier post-training, CaBM autonomously discovers high-quality, dataset-specific concepts. 
Our results across fine- and coarse-grained benchmarks demonstrate that CaBM achieves competitive accuracy while preserving interpretability without the constraints of external dictionaries or manual labeling. Our code is available at \url{https://github.com/bariscagliyan/CaptionBottleneckModels}.
\end{abstract}

\begin{figure*}[t]
  \centering
  \includegraphics[width=\textwidth]{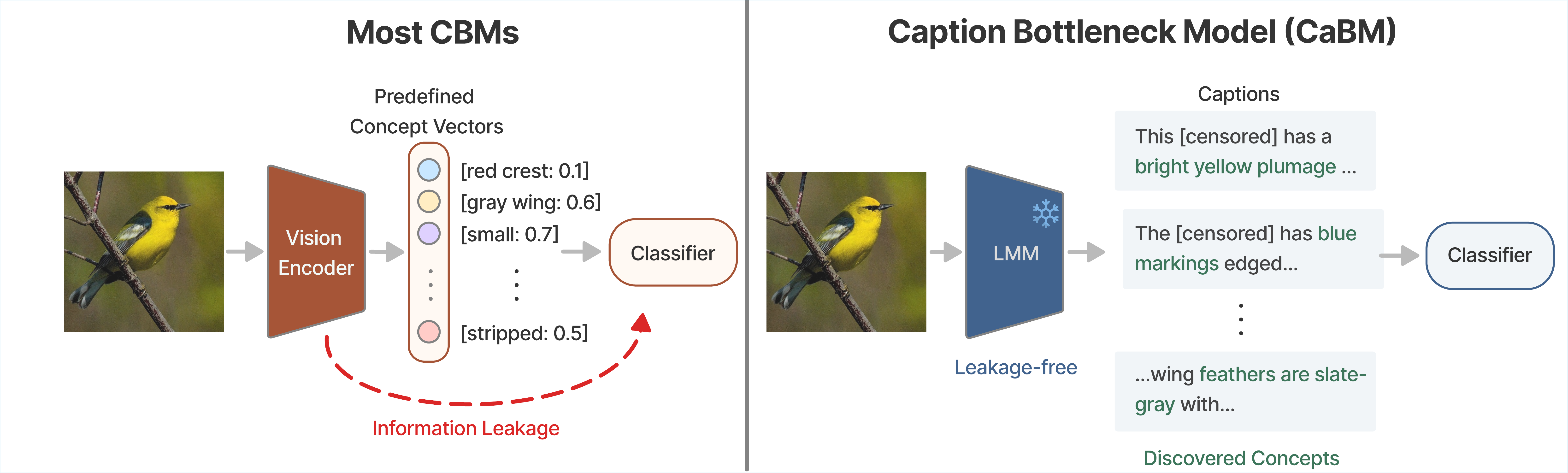}
  \caption{ CaBM performs recognition through a caption bottleneck:
  a frozen LMM produces captions, a text classifier predicts labels from these captions,
  and open-vocabulary concepts are extracted via post-hoc analysis.}
  \label{fig:teaser}
\end{figure*}

\section{Introduction}
\label{sec:intro}

Deep Neural Networks (DNNs) have achieved remarkable success across a wide range
of domains, including computer vision and natural language processing. Their
high accuracy and representational capacity have made them ubiquitous in
real-world systems. However, a fundamental limitation remains: the mechanisms
underlying their decisions are often opaque to human users. While
high-dimensional intermediate representations enable state-of-the-art
performance, they provide limited insight into \emph{how} and \emph{why} a
particular prediction is produced. This ``black-box'' behavior hinders reliable
deployment in high-stakes settings where interpretability, accountability, and user trust are essential
alongside predictive performance.

Concept Bottleneck Models (CBMs) provide a principled approach for 
 \emph{interpretable by design} predictors: they first map an input to a set of
human-interpretable concepts, and then predict the label from these concepts,
enabling inspection and test-time interventions on the concept layer
\cite{koh2020concept}. Subsequent works broadened the CBM paradigm in several
directions, including post-hoc explanation of pretrained models with concept
bottlenecks \cite{yuksekgonul2023posthoc} and interactive variants that query
humans for concept labels at inference time \cite{Chauhan2023InteractiveCBM}.
More recently, foundation models have been used to reduce the cost of building
and annotating concept interfaces by automatically proposing candidate
attributes and weakly grounding them with vision--language embeddings
\cite{Yang_2023_CVPR,Oikarinen2023LFCBM,Yan_2023_ICCV,Srivastava2024VLGCBM}. This trend has also motivated concept-based explanations for zero-shot
vision--language predictions \cite{ozdemir2026explaining}, alongside systematic
analyses of modern CBMs' design choices, evaluation pitfalls, and performance
trade-offs \cite{tapli2026rethinking}.

Despite rapid progress, current CBM pipelines still face two open problems:

\textbf{(i) Obtaining a concept set.}
A central practical bottleneck in developing CBMs is specifying a concept vocabulary that is both
\emph{human-meaningful} and \emph{task-sufficient}. Early CBMs relied on image-level concept annotations provided by domain experts \cite{koh2020concept}, which yields high-quality concept labels but is costly and hard to scale. 
To reduce supervision, recent methods
increasingly \emph{outsource concept definition} to foundation models. Typically, LLMs propose candidate concepts (often from class names), which are then grounded via zero-shot VLM alignment  (\eg using CLIP \cite{radford2021learning}) 
\cite{Liu_2025_CVPR,Oikarinen2023LFCBM,Yu_2025_CVPR}. 
Although scalable, this approach often generates noisy or uncertain labels that undermine explanation fidelity and intervention reliability \cite{Park_2025_NeurIPS}. Even when additional filtering
or selection is applied, these methods still begin from \emph{externally proposed}, typically static LLM/VLM-mediated concept sets that may be incomplete,
biased toward label semantics, or poorly grounded in the dataset’s actual visual evidence
\cite{Liu_2025_CVPR,Yang_2023_CVPR,Yan_2023_ICCV,Tan2024OpenCBM}.

\textbf{(ii) Information leakage undermining interpretability.} 
The presence of a concept layer  does not guarantee that
internal representations capture \emph{only} the intended concept
information. 
Multiple studies show that CBMs, particularly ``soft'' CBMs
trained jointly or sequentially, can encode task-relevant signals that bypass
the intended concept bottleneck layer, yielding high accuracy but unfaithful
explanations
\cite{Margeloiu2021DoCBMLearn,Mahinpei2021PromisesPitfalls,Havasi2022Leakage}.
This tension is often framed as an accuracy-faithfulness trade-off: 
passing \emph{continuous concept probabilities} boosts accuracy but creates a high-bandwidth channel for information leakage  that can corrupt concept
predictions and undermine interventions \cite{Havasi2022Leakage}. 
Conversely, independently training the concept and classification layers of CBMs on hard  (binarized) concepts makes them substantially more resilient to leakage, but significantly degrades their accuracy \cite{Havasi2022Leakage}.

We propose \textbf{CaBM} (\emph{Caption Bottleneck Models}), a new paradigm
that addresses both challenges by \emph{moving the bottleneck from structured
concept vectors to natural language} and \emph{enforcing independence} of the two stages (concept layer and classifier) \emph{by design} (\cref{fig:teaser}).
Instead of predicting a fixed concept set, CaBM first translates an image into
multiple \emph{diverse} and \emph{attribute-centric captions} using a \emph{frozen}
off-the-shelf Large Multimodal Model (LMM) guided by a structured prompt. 
We then train a
text classifier on these captions to predict the original image label, turning
recognition into language-space classification through a caption bottleneck.
Importantly, this two-stage design mirrors the leakage-resistance principle of
\emph{independent training of CBM with hard concepts}\cite{Havasi2022Leakage}: the downstream predictor cannot shape the
bottleneck through end-to-end gradients, and it never observes pixels -- only the
discrete text produced upstream. To further eliminate trivial shortcuts, we
apply deterministic \emph{taxonomy censoring} that removes explicit class names
and higher-level category terms from the captions, ensuring that prediction relies on
descriptive visual evidence rather than label-name cues.

Operating in free-form language space also enables \emph{concept discovery}: after training the text classifier, we identify class-discriminative phrases directly from its decision process and aggregate them into per-class concept vocabularies. These concepts are \emph{open-vocabulary by construction}, \emph{data- and model-induced}, and naturally interpretable because they correspond to literal spans in the generated captions rather than opaque latent variables.

Our work makes the following contributions:
(i) We introduce \textbf{CaBM}, a model-agnostic framework that decouples
    perception and recognition by performing classification \emph{entirely in
    language space} through a caption bottleneck.
(ii) We propose a practical \textbf{leakage-free caption pipeline} with
    structured multi-caption generation and deterministic taxonomy censoring,
    enabling robust caption-based supervision without label-name shortcuts.
(iii) We introduce an automated \textbf{open-vocabulary concept extraction} approach that recovers per-class concepts directly from the trained classifier’s behavior, eliminating predefined concept sets while yielding human-meaningful explanations.

\section{Related Work}
\label{sec:related-work}
\textit{Open-vocabulary CBMs and concept-set construction.}
A major line of work aims to reduce the reliance on expert concept annotations by
constructing concept sets automatically using foundation models. Approaches such as
LaBo \cite{Yang_2023_CVPR} and label-free CBMs \cite{Oikarinen2023LFCBM,Yan_2023_ICCV,Srivastava2024VLGCBM} query LLMs for candidate attributes and use CLIP-style
vision--language alignment to associate images with concept text, enabling scalable
concept supervision without concept-labeled datasets. Similarly, EZPC \cite{ozdemir2026explaining} projects CLIP's joint image–text embeddings into a predefined concept space via a learned linear projection, providing concept-level explanations of zero-shot predictions.
OpenCBM \cite{Tan2024OpenCBM} further relaxes the \emph{fixed} concepts by allowing users to specify
arbitrary textual concepts \emph{after} training via CLIP-space reconstruction, enabling
flexible add/remove/replace operations.
However, these pipelines still fundamentally operate over an \emph{externally specified}
concept set (LLM-proposed or user-provided text concepts) whose coverage and bias are
determined outside the target dataset; consequently, they can miss dataset-specific cues
or over-emphasize label semantics rather than grounded evidence.

\textit{Automated concept discovery.}
Rather than defining concepts \emph{a priori}, DN-CBM \cite{Rao2024DNCBM} inverts the usual workflow by
first discovering a sparse set of latent ``concepts'' by training a sparse auto-encoder on a pretrained backbone and
then naming them post-hoc by matching to a large vocabulary in a text-embedding space. This task-agnostic discovery improves reusability across datasets,
but the semantic interface remains mediated by the choice of an external vocabulary and
embedding space used for naming (\eg, nearest-neighbor matching), which can constrain
expressiveness and yield brittle or overly generic names. Extending this discovery line, DCBM~\cite{prasse2024dcbm} defines concepts as image regions
obtained from segmentation/detection foundation models, improving data efficiency while keeping
concepts \emph{visual} rather than textual, and OCB~\cite{steinmann2025object} couples a CBM
with a pretrained object-centric model to ground concepts at the object level. Both remain in the
image domain and rely on visual concepts.  HybridCBM \cite{Liu_2025_CVPR} addresses incomplete
predefined concept sets by complementing them with a learned \emph{dynamic} concept set
and a concept translator that maps learned vectors back to text.
While effective, HybridCBM still assumes a static set seeded by LLM-generated concepts
and learns additional concepts as continuous embeddings aligned to that set, so the
interpretation is ultimately tied to the external concept set and the translator’s
alignment quality. The same work also introduces CaptionCBM, a variant that replaces learned dynamic concepts with generated captions embedded as textual concepts; however, prediction still relies on CLIP image features at inference.

\textit{Language bottlenecks and caption-based explanations.}
A related direction replaces discrete concept lists with free-form textual explanations.
XBMs~\cite{yamaguchi2025explanation} train a vision encoder and explanation decoder
end-to-end to generate task-relevant natural-language explanations from image embeddings,
then predict the task label using a classifier that conditions on both the generated text
and image embeddings via cross-attention -- image features are thus present at every
stage of the pipeline, from explanation generation to final prediction.
In contrast, CaBM enforces a \emph{text-only} recognition channel by construction:
a frozen LMM produces (censored) captions and the downstream classifier never observes
pixels or image embeddings.

\textit{Positioning CaBM.}
We position CaBM as an alternative to both families: unlike open-vocabulary CBMs, it does not require
an externally proposed concept set to be specified \emph{before} or \emph{after} training;
instead, it performs recognition through a \emph{caption bottleneck} and discovers concepts
post-hoc from the trained text classifier’s evidence. Unlike concept discovery methods that
name latent vectors via dictionary lookup, CaBM’s concepts are literal spans drawn from the
model’s own caption evidence, yielding true open-vocabulary, dataset-specific phrases while
keeping the predictor strictly in the text domain.

\section{Method}
\label{sec:method}

\begin{figure}[t]
  \centering
  \includegraphics[width=\linewidth]{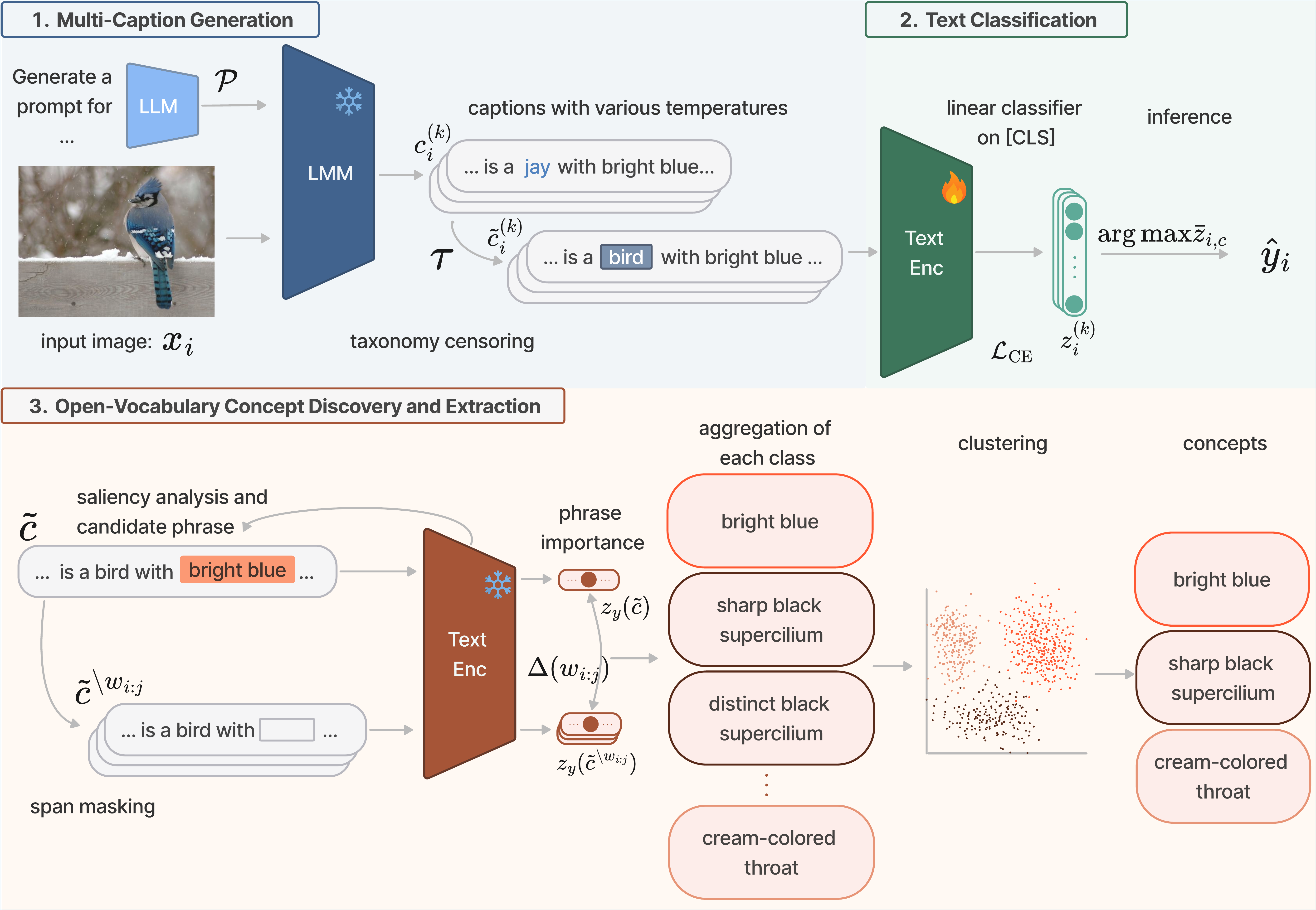}
  \caption{\textbf{Overview of CaBM.}
  \textbf{(1)} A frozen LMM generates $K$ diverse captions per image using a structured prompt $\mathcal{P}$
  and multiple decoding temperatures; captions are deterministically censored by $\tau$ to remove
  class names and taxonomy terms. \textbf{(2)} A text encoder predicts per-caption logits, which are averaged
  for image-level inference. \textbf{(3)} Concepts are obtained post-hoc by proposing salient spans
  (gradient$\times$embedding), scoring them with span masking (erasure), and semantically
  clustering phrases into compact per-class concept sets.}
  \label{fig:pipeline}
\end{figure}

CaBM decouples perception from recognition by translating images into text and performing
classification entirely on text. 
Our pipeline  (\cref{fig:pipeline}) has three stages:
(1)~\textbf{multi-caption generation} with taxonomy censoring,
(2)~\textbf{text classification} on the resulting caption dataset,
and (3)~\textbf{open-vocabulary concept discovery and extraction} from the trained classifier
and captions. 

\subsection{Multi-Caption Generation}
\label{sec:multi-caption-generation}
Given a labeled image dataset $\mathcal{D}=\{(\mathbf{x}_i,y_i)\}_{i=1}^{N}$ with
$y_i\in\{1,\dots,C\}$, we generate a set of diverse textual descriptions for each
image using a Large Multimodal Model (LMM). For each image
$\mathbf{x}_i$, we sample $K$ captions
$\mathcal{C}_i=\{c_i^{(k)}\}_{k=1}^{K}$:
\begin{equation}
c_i^{(k)} \sim p_{\theta}\!\left(\,\cdot \mid \mathbf{x}_i, \mathcal{P}; T^{(k)}\right),
\qquad k=1,\dots,K,
\end{equation}
where $\mathcal{P}$ is a fixed structured prompt and $T^{(k)}$ denotes the
sampling temperature.

\textit{Structured prompting.}
To make captions discriminative across diverse datasets, we construct a single
prompt template $\mathcal{P}$ using an LLM-assisted design step: given only the
dataset domain and the nature of class granularity in the dataset (coarse or fine-grained), the LLM proposes an attribute checklist (\eg, color, shape, texture/pattern, part-level cues, and spatial arrangement) and a constrained output format. The resulting prompt is frozen and used for all images. Importantly, $\mathcal{P}$ explicitly discourages taxonomic naming and encourages grounded descriptions.

\textit{Temperature-driven diversity.}
We generate $K$ complementary captions per image by keeping $\mathcal{P}$
fixed and varying the sampling temperature.
Lower temperatures favor concise, prototypical descriptions, while higher
temperatures promote lexical and compositional diversity, yielding multiple
textual ``views'' of the same image.

\textit{Decoding and normalization.}
We use standard stochastic decoding with conservative controls to stabilize
generation and reduce degenerate repetition, and apply lightweight text
normalization.

\textit{Taxonomy censoring.}
Even with constrained prompting, LMM outputs may contain label leakage via exact
class names or higher-level taxonomic/group terms. We therefore apply a
deterministic censoring function $\tau(\cdot)$ to each caption:
\begin{equation}
\tilde{c}_i^{(k)} = \tau\!\left(c_i^{(k)}\right),
\end{equation}
which replaces any matched class string and any matched group term with a dataset-generic referent (\eg, ``the object'').
This ensures downstream classification cannot rely on trivial lexical cues and
must instead use descriptive visual attributes.

\subsection{Text Classification}
\label{sec:text-classifier}
Given $K$ censored captions per image
$\mathcal{C}_i$ (\cref{sec:multi-caption-generation}),
we learn a caption-to-label classifier and predict image labels by aggregating
caption-level predictions. To prevent leakage between training and validation splits, we partition the data
\emph{at the image level} with class stratification and place all $K$ captions of
each image in the same split.

\textit{Model.}
Let $\mathrm{Enc}_{\phi}$ be a pretrained transformer encoder. A caption is
tokenized and truncated to a maximum length $L$, producing a
sequence of $T\leq L$ tokens. The encoder outputs
$\mathbf{H}=\mathrm{Enc}_{\phi}(\tilde{c})\in\mathbb{R}^{T\times d}$.
We use the final-layer \texttt{[CLS]} representation $\mathbf{h}\in\mathbb{R}^{d}$
and apply a linear classifier:
\begin{equation}
\mathbf{z}= \mathbf{W}\mathbf{h}+\mathbf{b}, \qquad
p_{\phi}(y \mid \tilde{c})=\mathrm{softmax}(\mathbf{z}),
\end{equation}
where $\mathbf{W}\in\mathbb{R}^{C\times d}$ and $\mathbf{b}\in\mathbb{R}^{C}$.

\textit{Training objective.}
We fine-tune $\mathrm{Enc}_{\phi}$ and the linear head by minimizing cross-entropy over
caption samples:
\begin{equation}
\mathcal{L}_{\text{CE}} = - \log p_{\phi}(y \mid \tilde{c}).
\end{equation}
Optimization details are provided in \cref{sec:experimental-setup}.

\textit{Multi-caption inference (logit voting).}
At test time, we compute logits $\mathbf{z}_i^{(k)}$ for each caption and predict
at the image level by averaging logits across captions:
\begin{equation}
\bar{\mathbf{z}}_i=\frac{1}{K}\sum_{k=1}^{K}\mathbf{z}_i^{(k)}, \qquad
\hat{y}_i=\underset{c\in\{1,\ldots,C\}}{\arg\max} \bar{z}_{i,c}.
\label{eq:avg_vote}
\end{equation}
This aggregation acts as a textual multi-view ensemble and reduces sensitivity
to any single caption.

\subsection{Open-Vocabulary Concept Discovery and Extraction}
\label{sec:open-vocab-concept-extraction}
A central design choice in CaBM is that classification is carried out entirely in the
text space, enabling the model's evidence to be decomposed into meaningful
phrases. We propose an automated, open-vocabulary concept extraction pipeline
that (i) proposes candidate phrases from correctly classified samples using
gradient-based attribution, (ii) rescores them via span-masking (erasure)
to obtain faithful importance scores, and (iii) consolidates and ranks them into
compact, per-class concept vocabularies through semantic clustering.

\textit{Correctness filtering.} To ensure that extracted phrases reflect features genuinely used for robust discrimination, we restrict extraction to correctly classified training images.

\textit{Gradient $\times$ embedding attribution.} For each caption $\tilde{c}_i^{(k)}$ of a retained image, we compute token-level
attribution scores using gradient $\times$ embedding \cite{ancona2018towards}.
Let $\mathbf{e}_t$ denote the embedding-layer representation of token $t$, and
let $z_{y_i}$ be the logit for the target class (ground-truth label during
extraction). We define the raw saliency of token $t$ as
\begin{equation}
s_t \;=\; \left\lVert \frac{\partial z_{y_i}}{\partial \mathbf{e}_t}
\odot \mathbf{e}_t \right\rVert_1,
\end{equation}
where $\odot$ is the Hadamard product. We zero the saliency of special tokens and normalize $\{s_t\}$ to sum to one over non-padding tokens, and sum subword scores to obtain word-level scores. 

\textit{Candidate phrase proposal (saliency n-grams).}
Given the word-level saliency map, we propose candidate phrases using
variable-length $n$-grams. For each span of words $w_{i:j}$, we
compute a lightweight \emph{proposal score} as the sum of its word saliencies.
We remove generic filler terms (\eg, ``image'', ``appears'') and edge stopwords,
and select the top $P$ candidates per caption using greedy non-maximum
suppression with a maximum of 50\% span overlap. This yields up to $KP$
candidate phrases per image.

\textit{Span-masking (erasure) scoring.}
Gradient attribution is used only to propose spans; \emph{phrase importance}
is computed via an erasure test that measures how much masking a
candidate span decreases the target-class logit \cite{li2016erasure}. For a
candidate phrase $w_{i:j}$ occurring in caption $\tilde{c}$, let
$\tilde{c}^{\setminus w_{i:j}}$ denote the caption where the corresponding token
span is erased (by dropping them via padding and zeroing the attention mask). We
define the \emph{phrase importance} as the (clipped) logit drop:
\begin{equation}
\Delta(w_{i:j}) \;=\; \max\Bigl(0,\; z_{y}(\tilde{c}) - z_{y}(\tilde{c}^{\setminus w_{i:j}})\Bigr).
\end{equation}

\textit{Semantic clustering via encoder embeddings.}
We treat extracted phrases as short texts and map them to a semantic embedding
space using the trained CaBM encoder's \texttt{[CLS]}-style pooled
representation, which is standard in transformer encoders for classification
and sentence-level representations \cite{devlin2018bert}. Within each class, we
cluster phrase embeddings with HDBSCAN \cite{mcinnes2017hdbscan}, which is
well-suited to discovering variable-density groups while marking outliers as
noise. To reduce redundancy, we
apply a lightweight post-processing merge based on centroid similarity and
string-level normalization (substring/word-overlap), so that near-duplicate
surface forms are consolidated into a single concept. Each cluster
$\mathcal{G}$ is summarized by a representative phrase:
\begin{equation}
p^{\ast}(\mathcal{G}) \;=\; \arg\max_{p \in \mathcal{G}}
\cos\!\bigl(\mathbf{e}(p), \boldsymbol{\mu}_{\mathcal{G}}\bigr).
\end{equation}
where $\mathbf{e}(p)$ is the L2-normalized phrase embedding and $\boldsymbol{\mu}_{\mathcal{G}}$
is the (L2-normalized) cluster centroid. Finally, we assign each concept a
\emph{concept importance} score by aggregating phrase-level erasure evidence:
\begin{equation}
S(\mathcal{G}) \;=\; \sum_{m=1}^{|\mathcal{G}|} \Delta(p_m),
\end{equation}
and rank concepts in descending order of $S(\mathcal{G})$.

\section{Experiments}
\label{sec:experiments}

We comprehensively evaluate CaBM across six coarse- and fine-grained recognition benchmarks to validate its effectiveness, interpretability, and predictive performance. Our evaluation is structured around three core pillars. First, we assess the quality of our open-vocabulary concept discovery both quantitatively and qualitatively (\cref{sec:concept-quant,sec:concept-qual}), demonstrating that CaBM extracts semantically valid and highly visually grounded concepts compared to state-of-the-art baselines. Second, we evaluate the faithfulness of our model through test-time human interventions (\cref{sec:intervention}), illustrating that modifying the caption bottleneck directly and predictably alters classification outcomes. Finally, we benchmark overall classification accuracy (\cref{sec:accuracy}), showing that despite operating strictly in the text domain during inference, CaBM achieves highly competitive predictive performance against traditional image-domain CBMs.

\subsection{Experimental Setup}
\label{sec:experimental-setup}

\textit{Datasets.}
We evaluate CaBM on six benchmarks: CIFAR-10, CIFAR-100~\cite{krizhevsky2009learning}, CUB-200-2011~\cite{wah2011cub}, Food-101~\cite{bossard2014food101}, Flowers-102~\cite{nilsback2008flowers102}, ImageNet-1K~\cite{deng2009imagenet}.
For datasets without an official validation split, we create a stratified
validation set by holding out 10\% of training images; all $K$ captions of an
image remain in the same split.

\textit{Caption generation.}
We generate $K{=}5$ captions per image using structured prompting
(\cref{sec:multi-caption-generation}) at temperatures
$\{0.7,0.9,1.1,1.3,1.5\}$. For ImageNet-1K, we use $K{=}3$ captions
from Qwen3-VL-8B-Instruct at temperatures $\{0.0,0.3,0.6\}$
for scalability.
For all other datasets we employ a frozen LMM,
Qwen3-VL-2B-Instruct~\cite{bai2025qwen3vl},
and apply deterministic taxonomy censoring ($\tau$).

\textit{Text classifier.}
We use RoBERTa-base~\cite{liu2019roberta} for all datasets and train a linear
classification head on the final-layer \texttt{[CLS]} representation, as
described in \cref{sec:text-classifier}. We fine-tune with AdamW
($\text{wd}{=}0.05$), a cosine learning-rate schedule with 6\% warm-up, label
smoothing ($\epsilon{=}0.1$), effective batch size 48,
maximum sequence length 512, and early stopping
with patience 6 for up to 30 epochs. We additionally use layer-wise
learning-rate decay with factor 0.85, freeze the embeddings and the two lowest
encoder layers, and apply light caption-level augmentation during training only
(word dropout $p{=}0.1$ and short span dropout $p{=}0.03$, applied with
probability 0.3). No augmentation is used at validation or test time.

At test time, we predict each image's label by averaging logits across its $K$
captions (\cref{eq:avg_vote}).
Details about compute and running time can be found in the supplementary material. 

\subsection{Open-Vocabulary Concept Discovery: Quantitative Evaluation}
\label{sec:concept-quant}

\paragraph{Concept quality.}
Beyond qualitative inspection, we evaluate the semantic quality of per-class
concept sets using the metrics introduced in HybridCBM~\cite{Liu_2025_CVPR}.
Given a concept set that provides textual concepts for each class
$k \in \{1,\dots,C\}$, we embed both concept phrases and class names with a CLIP
text encoder~\cite{radford2021learning}, and report:

\textbf{(i) Purity}: the cosine similarity between the class-wise mean concept embedding (L2-normalized) and the CLIP text embedding of the corresponding class name, averaged over classes.

\textbf{(ii) Separation}: inter-class distinctiveness measured as the average pairwise cosine distance between class-mean concept embeddings; larger distances indicate more differentiated concept sets.

\textbf{(iii) Semantics}: an LLM-based validity rate (GPT-3.5) obtained from binary yes/no judgments of whether a concept phrase is associated with its intended class, reflecting interpretability and class relevance.

\cref{tab:concept-quality-cub} reports results on
CUB-200. CaBM attains high semantic validity with competitive purity and
improved separation. Importantly, CaBM concepts are \emph{discovered}, rather than selected from an
externally proposed, \emph{LLM-generated} vocabulary as in VLG-CBM. Despite this,
CaBM surpasses VLG-CBM in Semantics and Separation, while matching purity.

HybridCBM shows the highest Separation, but this is partly driven by its orthogonality
regularization, which increases inter-class cosine distance by construction. In our
evaluation, this comes with lower Purity and lower Semantics.
CaBM instead yields image-grounded concepts, preserving strong Purity while
substantially improving semantic validity at moderate Separation.

\begin{table}[t]
  \centering
  \caption{\textbf{Quantitative comparison of concept sets on CUB200.} Bold values indicate the best performance.}
  \label{tab:concept-quality-cub}
  \begin{tabular}{lccc}
    \toprule
    Method (CUB) & Purity & Separation & Semantics \\
    \midrule
    VLG-CBM (NeurIPS'24)~\cite{Srivastava2024VLGCBM} & \textbf{0.52} & 0.07 & 0.86 \\
    HybridCBM (CVPR'25)~\cite{Liu_2025_CVPR} & 0.40 & \textbf{0.80} & 0.46 \\
     CaBM & \textbf{0.52} & 0.16 & \textbf{0.92} \\
    \bottomrule
  \end{tabular}
\end{table}

\begin{table}[t]
  \centering
  \caption{\textbf{Downstream utility under NEC-controlled VLG-CBM.}
  We keep the VLG-CBM evaluation protocol fixed and replace only the concept list:
  the original LLM-generated concepts vs.\ CaBM-discovered concepts.
  Top-1 accuracy is reported as ANEC-5 (NEC{=}5) and ANEC-avg (NEC $\in \{5,10,15,20,25,30\}$) following~\cite{Srivastava2024VLGCBM}.}
  \label{tab:anec-conceptset}
  \small
  \setlength{\tabcolsep}{4.5pt}
  \begin{tabular}{lcccccc}
    \toprule
    & \multicolumn{2}{c}{C10} & \multicolumn{2}{c}{C100} & \multicolumn{2}{c}{CUB} \\
    \cmidrule(lr){2-3}\cmidrule(lr){4-5}\cmidrule(lr){6-7}
     & ANEC-5 & ANEC-avg & ANEC-5 & ANEC-avg & ANEC-5 & ANEC-avg \\
    \midrule
    VLG-CBM & 88.55 & 88.63 & 65.73 & 66.48 & 75.79 & 75.82 \\
    CaBM     & \textbf{88.87} & \textbf{88.97} & \textbf{66.13} & \textbf{66.66} & \textbf{76.25} & \textbf{76.34} \\
    \bottomrule
  \end{tabular}
\end{table}

\textit{Downstream utility under controlled effective concepts.}
Complementary to CLIP-based concept-quality metrics (Purity/Separation) and LLM Semantics,
we test whether the discovered concepts form a discriminative vocabulary under the
NEC-controlled evaluation of VLG-CBM~\cite{Srivastava2024VLGCBM}.
We keep the VLG-CBM evaluation pipeline fixed and swap \emph{only} the concept set
(LLM-generated vs.\ CaBM-discovered). \cref{tab:anec-conceptset} shows consistent gains on CIFAR-10,
CIFAR-100, and CUB, indicating that CaBM concepts capture decision-relevant evidence even when 
constrained to a small, inspectable set of effective concepts.

\subsection{Concept Discovery: Qualitative Comparisons}
\label{sec:concept-qual}

\begin{figure}[t]
  \centering
  \includegraphics[width=\linewidth]{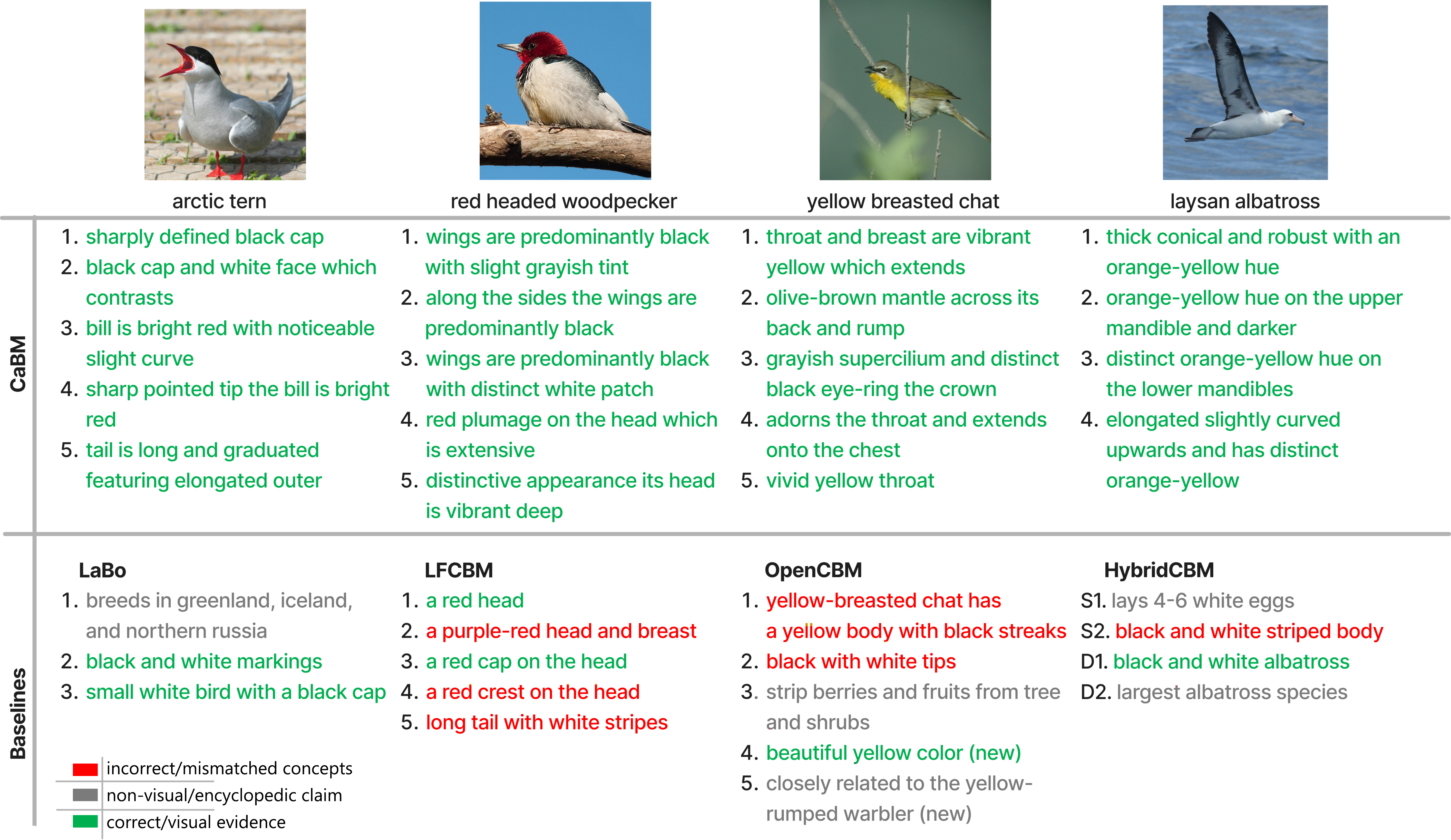}
\caption{\textbf{Qualitative concept comparison across methods on CUB200.}
For each class, we list top-ranked concepts from CaBM and baselines.
Concepts are manually tagged on the shown instance:
OpenCBM marks \emph{(new)} expansions; HybridCBM reports \emph{Static} (S) and \emph{Dynamic} (D). For a strictly fair comparison, we used  the specific test images originally selected by the baseline papers. }
  \label{fig:qualitative-comparison}
\end{figure}

\paragraph{Cross-method concept inspection.}
\cref{fig:qualitative-comparison} contrasts the types of concepts produced by each method. \emph{For fairness, both the image instances and the listed concepts are taken directly
from the experiments of the corresponding baseline papers; they are not selected or curated by us.}
We run CaBM on these identical instances to enable a \emph{like-for-like comparison.}
Across examples, CaBM surfaces attribute-centric phrases that correspond to directly inspectable
visual cues (\eg, part color/shape and distinctive markings), yielding compact, instance-level
explanations.

In comparison, baselines that rely on externally proposed concept sets or post-hoc translation
exhibit a mixture of visually grounded evidence and statements that are either mismatched to the
shown instance (red) or not visually verifiable (gray). For example, LaBo\cite{Yang_2023_CVPR} may
include geographic breeding information; OpenCBM’s\cite{Tan2024OpenCBM} expansion can increase coverage but also introduces
relational/encyclopedic concepts (\eg, diet or relatedness); and HybridCBM’s\cite{Liu_2025_CVPR} translated concepts can
be diverse yet sometimes describe non-visual properties or mismatched attributes. LF-CBM\cite{Oikarinen2023LFCBM} often produces
semantically overlapping variants around a small number of cues, leading to redundancy. Overall, the
tagging highlights that CaBM’s concepts are typically concise and visually grounded with reduced
redundancy relative to the compared baselines.

\begin{figure}[t]
  \centering
  \includegraphics[width=\linewidth]{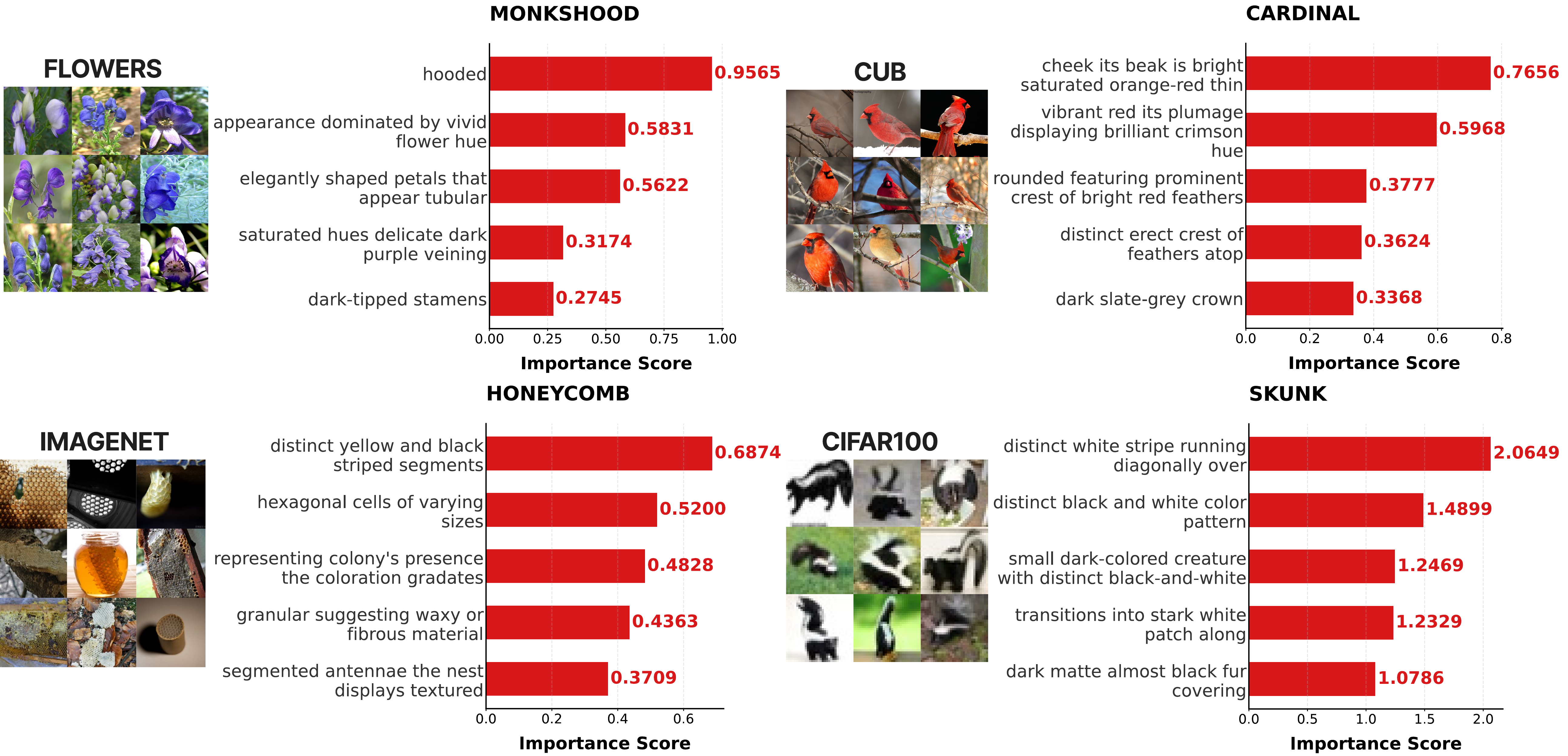}
  \caption{\textbf{Concept importance across images.} For four datasets, we select one
  random class and sample nine random test images. For each class, we show the top-5
  concepts and their importance scores (higher indicates stronger influence under the
  ranking criterion in \cref{sec:open-vocab-concept-extraction}).}
  \label{fig:concept-activation}
\end{figure}

\textit{Concept importance across instances.}
To assess whether the discovered concepts provide consistent, instance-level
evidence rather than isolated artifacts, we visualize concept importance across
multiple images from the same class. \cref{fig:concept-activation} shows four
randomly selected classes (one per dataset) and nine randomly sampled test images
per class, together with the top-5 concepts and their importance scores (as used
for ranking in \cref{sec:open-vocab-concept-extraction}). Across datasets, the
highest-ranked concepts correspond to recurring, visually diagnostic attributes
that reappear across diverse views and backgrounds within the class, while lower
ranked concepts capture complementary cues. This supports that the concept sets
are compact and reusable across instances, enabling inspection of which evidence
is most influential for a given prediction.

\subsection{Faithfulness via Human Intervention}
\label{sec:intervention}

A key advantage of concept bottleneck models is \emph{human intervention}: users can
edit concepts and observe predictable changes in the output. We probe this property
in CaBM by intervening directly at the caption bottleneck. For each dataset, we form
an \emph{intervention subset} consisting of test images from the 20 classes with the
highest misclassification frequency. For each test instance, we inject the top-ranked concept from its ground-truth class to the caption and re-evaluate the same frozen text classifier.
If the discovered concepts function as actionable evidence, such minimal edits should
systematically improve accuracy.

\begin{figure}[t]
  \centering
  \includegraphics[width=\linewidth]{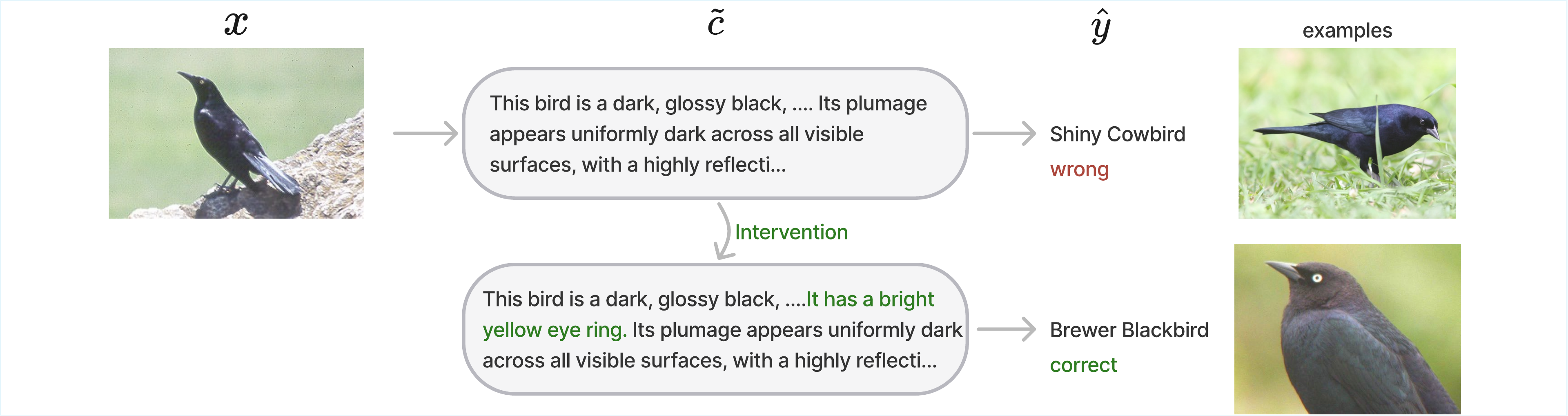}
\caption{\textbf{Human intervention example.}
Prepending one concept to the caption flips the prediction under the same frozen classifier.}
  \label{fig:intervention-example}
\end{figure}

\textit{Results.}
\cref{tab:intervention-acc} reports accuracy on the intervention subset. Concept
injection yields consistent gains across datasets ($+12.2$ to $+13.8$ points).
\cref{fig:intervention-example} shows a representative case where a single injected
concept flips an incorrect prediction to the correct class.

\begin{table}[t]
  \centering
\caption{\textbf{Human intervention via concept injection.}
Accuracy on the intervention subset (test images from the 20 most misclassified classes).
We inject one top-ranked concept of the ground-truth class to each caption and re-evaluate the frozen classifier.}
  \label{tab:intervention-acc}
  \small
  \setlength{\tabcolsep}{5pt}
  \begin{tabular}{lccc}
    \toprule
    Dataset & Baseline Acc. (\%) & Injected Acc. (\%) & $\Delta$ \\
    \midrule
    Flowers-102 & 60.6 & 72.8 & \textcolor{green!50!black}{+12.2} \\
    CIFAR-100   & 62.3 & 75.2 & \textcolor{green!50!black}{+13.0} \\
    CUB-200     & 41.7 & 55.5 & \textcolor{green!50!black}{+13.8} \\
    \bottomrule
  \end{tabular}
\end{table}

\subsection{Classification Performance}
\label{sec:accuracy}
\cref{tab:accuracy_table} reports Top-1 test accuracy across six benchmarks.
Despite operating \emph{entirely} in the text domain at inference time, CaBM achieves competitive performance on both coarse-
and fine-grained recognition tasks.

We compare against representative CBM baselines:
PCBM~\cite{yuksekgonul2023posthoc}, LF-CBM~\cite{Oikarinen2023LFCBM}, LaBo~\cite{Yang_2023_CVPR}, VLG-CBM~\cite{Srivastava2024VLGCBM} and PS-CBM~\cite{zhao2025partially} which operate in the \emph{image} domain (typically via a CLIP-style image encoder and a concept bottleneck). We further include XBM~\cite{yamaguchi2025explanation}, which uses a \emph{multimodal} classifier over image and generated explanation text, retaining visual features at inference—fundamentally different from CaBM’s strictly text-only pathway.

\begin{table}[t]
  \centering
    \caption{\textbf{Classification accuracy (\%) comparison.}
    Methods are grouped by bottleneck/protocol. CaBM predicts from LMM captions only,
    whereas other CBMs use visual backbones or image-derived concept representations.
    Baseline results are taken from the cited papers; LF-CBM, LaBo, VLG-CBM, and
    PS-CBM follow~\cite{zhao2025partially}. \textbf{--}: not reported.}
  \label{tab:accuracy_table}
  \small
  \setlength{\tabcolsep}{3.0pt}
  \begin{tabular}{@{}lcccccc@{}}
    \toprule
    Method & C10 & C100 & CUB & Food & Flowers & IN-1K \\
    \midrule
    \multicolumn{7}{@{}l}{\textit{CBMs with soft concepts (CNN)}} \\
    PCBM (ICLR'23)~\cite{yuksekgonul2023posthoc}
      & 77.70 & 52.00 & 58.80 & --    & --    & -- \\
    LF-CBM (ICLR'23)~\cite{Oikarinen2023LFCBM}
      & 87.30 & 68.80 & 58.60 & 77.70 & 94.40 & 67.50 \\
    LaBo (CVPR'23)~\cite{Yang_2023_CVPR}
      & 87.50 & 68.10 & 66.80 & 82.20 & 96.60 & 72.10 \\
    VLG-CBM (NeurIPS'24)~\cite{Srivastava2024VLGCBM}
      & 89.60 & 68.30 & 68.00 & 81.60 & 97.10 & 65.70 \\
    PS-CBM (AAAI'26)~\cite{zhao2025partially}
      & 89.80 & 72.10 & 70.10 & 83.00 & 97.90 & 74.00 \\
    \midrule
    \multicolumn{7}{@{}l}{\textit{CBMs with soft concepts (ViT)}} \\
    HybridCBM (CVPR'25)~\cite{Liu_2025_CVPR}
      & 97.93 & 86.22 & 84.25 & 92.62 & 99.23 & 83.67 \\
    CBM-Suite (CVPR'26)~\cite{tapli2026rethinking}
      & --    & 92.50 & 86.73 & --    & --    & 81.56 \\
    \midrule
    \multicolumn{7}{@{}l}{\textit{Explanation bottleneck (text\,+\,image classifier)}} \\
    XBM (AAAI'25)~\cite{yamaguchi2025explanation}
      & --    & --    & 80.99 & --    & --    & 67.83 \\
    \midrule
    \multicolumn{7}{@{}l}{\textit{Strictly text bottleneck (text-only classifier)}} \\
    CaBM
      & 96.42 & 81.14 & 76.92 & 93.68 & 84.23 & 73.90 \\
    \bottomrule
  \end{tabular}
\end{table}

\textit{Discussion.}
\cref{tab:accuracy_table} shows that CaBM attains competitive Top-1 accuracy
under a strict \emph{text-only} inference constraint. Direct accuracy comparison across CBMs is inherently limited because the methods differ in their visual backbones, concept set construction, and whether the classifier has access to image features. The purpose of this table is therefore not to claim that CaBM is the strongest vision classifier, but to show that a strictly text-only caption bottleneck remains competitive while providing faithful and inspectable concept-based explanations. Stronger vision-backbone CBMs such as HybridCBM and CBM-Suite achieve higher accuracy on several datasets, whereas CaBM trades some visual information for a leakage-resistant, open-vocabulary bottleneck whose concepts are mined directly from the captions.

\section{Conclusion, Discussion \& Limitations}
\label{sec:conclusion}
In this work, we introduced Caption Bottleneck Models (CaBM), an interpretable-by-design framework that overcomes the reliance on predefined concept vocabularies and the vulnerability to information leakage inherent in traditional Concept Bottleneck Models (CBMs). By shifting the information bottleneck entirely into the natural language domain, CaBM strictly decouples perception from recognition. Using a frozen Large Multimodal Model (LMM) we generate diverse, taxonomy-censored captions, from which an independent text classifier predicts the target label. This discrete text interface ensures a leakage-free architecture by construction. Furthermore, our automated extraction pipeline successfully discovers dataset-specific, open-vocabulary concepts via gradient attributions and erasure signals. Evaluations across diverse benchmarks confirm that CaBM achieves competitive accuracy while yielding highly faithful, human-interpretable explanations that support targeted test-time interventions.

The evolution of CBMs reflects a broader trajectory toward scalable interpretability. Early CBMs relied on costly manual annotations, while subsequent generations leveraged LLMs to propose concepts and Vision-Language Models (VLMs) to ground them. However, these methods remained constrained by externally proposed, static concept sets.  The CaBM paradigm represents the next logical step, enabled by recent advancements in powerful LMMs (\eg, Qwen-VL) and robust text encoders (\eg, RoBERTa). By harnessing these models to generate and classify rich, accurate visual descriptions, CaBM eliminates the need for predefined concept sets altogether, allowing the data's true visual evidence to drive concept discovery.

\textit{Limitations.} While CaBM offers robust, leakage-free interpretability, it introduces a distinct architectural trade-off. A primary goal of certain post-hoc CBM variants (\eg, PCBM) is the ability to retroactively convert an arbitrary, pre-trained black-box vision encoder into an explainable model. Because CaBM is inherently interpretable-by-design, bypassing traditional vision encoders in favor of a specialized LMM-to-text pipeline, it cannot be attached to existing black-box models to explain their internal representations. Consequently, CaBM is best suited for scenarios where deploying a novel, transparent pipeline is preferred over auditing a legacy vision backbone.

\section*{Acknowledgements}
The numerical calculations reported in this paper were fully performed at TUBITAK ULAKBIM, High Performance and Grid Computing Center (TRUBA resources).
We also gratefully acknowledge the computational resources provided
by METU Center for Robotics and Artificial Intelligence
(METU-ROMER). The presentation of this work at ECCV 2026 was supported by TÜBİTAK under the 2224-A program.

\bibliographystyle{splncs04}
\bibliography{main}
\end{document}